\documentclass[11pt,a4paper]{article}
\usepackage[final]{acl}
\usepackage{latexsym}

\usepackage{booktabs}
\usepackage{graphicx}
\usepackage{array}
\usepackage{arydshln}
\usepackage{enumitem}
\usepackage{microtype}
\usepackage{xcolor}
\usepackage{amsmath,amssymb}
\usepackage{enumitem}
\usepackage{microtype}
\usepackage{pifont}
\newcommand{\cmark}{\ding{51}}%
\newcommand{\xmark}{\ding{55}}%

\date{}
\definecolor{darkgreen}{rgb}{0,0.5,0}
\definecolor{lightblue}{rgb}{0.7,0.9,1}
\definecolor{lblue}{rgb}{0.25,0.5,1}

\def\ODdel#1{\bgroup\markoverwith{\textcolor{cyan!89!yellow!80!black!100}{\rule[0.4ex]{2pt}{3pt}}}\ULon{#1}}
\def\VDdel#1{\bgroup\markoverwith{\textcolor{green!90!red!55}{\rule[0.4ex]{2pt}{3pt}}}\ULon{#1}}

\def\TNdel#1{\bgroup\markoverwith{\textcolor{blue!80!black}{\rule[0.4ex]{2pt}{3pt}}}\ULon{#1}}

\def\DSdel#1{\bgroup\markoverwith{\textcolor{blue!80!yellow!80!black!90}{\rule[0.4ex]{2pt}{3pt}}}\ULon{#1}}

\def\ZKdel#1{\bgroup\markoverwith{\textcolor{green!60!black!100}{\rule[0.4ex]{2pt}{3pt}}}\ULon{#1}}

\newcommand{\pz}{\phantom{0}}

\title{Learning Interpretable Latent Dialogue Actions With Less Supervision}

\usepackage[normalem]{ulem}

\usepackage{times}
\usepackage[T1]{fontenc}

\usepackage[utf8]{inputenc}
\usepackage{multirow}
\usepackage{microtype}

\author{Vojtěch Hudeček and Ondřej Dušek \\
\texttt{hudecek@ufal.mff.cuni.cz}, \texttt{odusek@ufal.mff.cuni.cz}\\
Charles University, Faculty of Mathematics and Physics\\Malostranské náměstí 25, 118 00 Prague, Czechia
}

\begin{document}
\maketitle

\begin{abstract}
We present a novel architecture for explainable modeling of task-oriented dialogues with discrete latent variables to represent dialogue actions. Our model is based on variational recurrent neural networks (VRNN) and requires no explicit annotation of semantic information. Unlike previous works, our approach models the system and user turns separately and performs database query modeling, which makes the model applicable to task-oriented dialogues while producing easily interpretable action latent variables.
We show that our model outperforms previous approaches with less supervision in terms of perplexity and BLEU on three datasets, and we propose a way to measure dialogue success without the need for expert annotation.
Finally, we propose a novel way to explain semantics of the latent variables with respect to system actions.

\end{abstract}

\section{Introduction}
While supervised neural dialogue modeling is a very active research topic 
\cite{wen2016network,lei2018,peng2021soloist}, it requires a significant amount of work to obtain turn-level labels, usually with dialogue state annotation.
We argue that in many real-world cases, it is very expensive to obtain the necessary labels or even to design an appropriate annotation schema.
Consider a call center with various dialogues that has a lot of transcripts available, including the corresponding API queries, but has no capacity to label them.
This motivates our research of approaches that minimize the need for expert annotation.

While most recent research focuses on pretrained language models (PLMs) and reaches state-of-the-art performance in standard supervised 
\cite{peng2021soloist,zhang2020dialogpt} or even few-shot \cite{peng2020few,wu2020-todbert} settings, these models still require full supervision.
Furthermore, they lack the potential to interpret the model decisions.
Some recent works try to address PLM interpretability with some success \cite{lin-etal-2019-open, stevens-su-2021-investigation},
but still face considerable
difficulties due to PLMs' huge number of parameters and their structure.
On the other hand, dialogue models using latent variables are able to infer interpretable attributes from unlabeled data \cite{wen2017latent,zhao-etal-2019-rethinking}.
These models are mostly trained using variational autoencoders \cite[VAE; ][]{kingma2013auto,serban2016hierarchical}.
Improvements with discrete variables \cite{zhao2018unsupervised,shi2019unsupervised} offer better interpretability, but the approaches are not directly applicable to task-oriented response generation as no distinction between the system and user roles is made, and database access or goal fulfillment are not considered;
\begin{figure*}[t]
    \centering
    \includegraphics[width=0.9\textwidth]{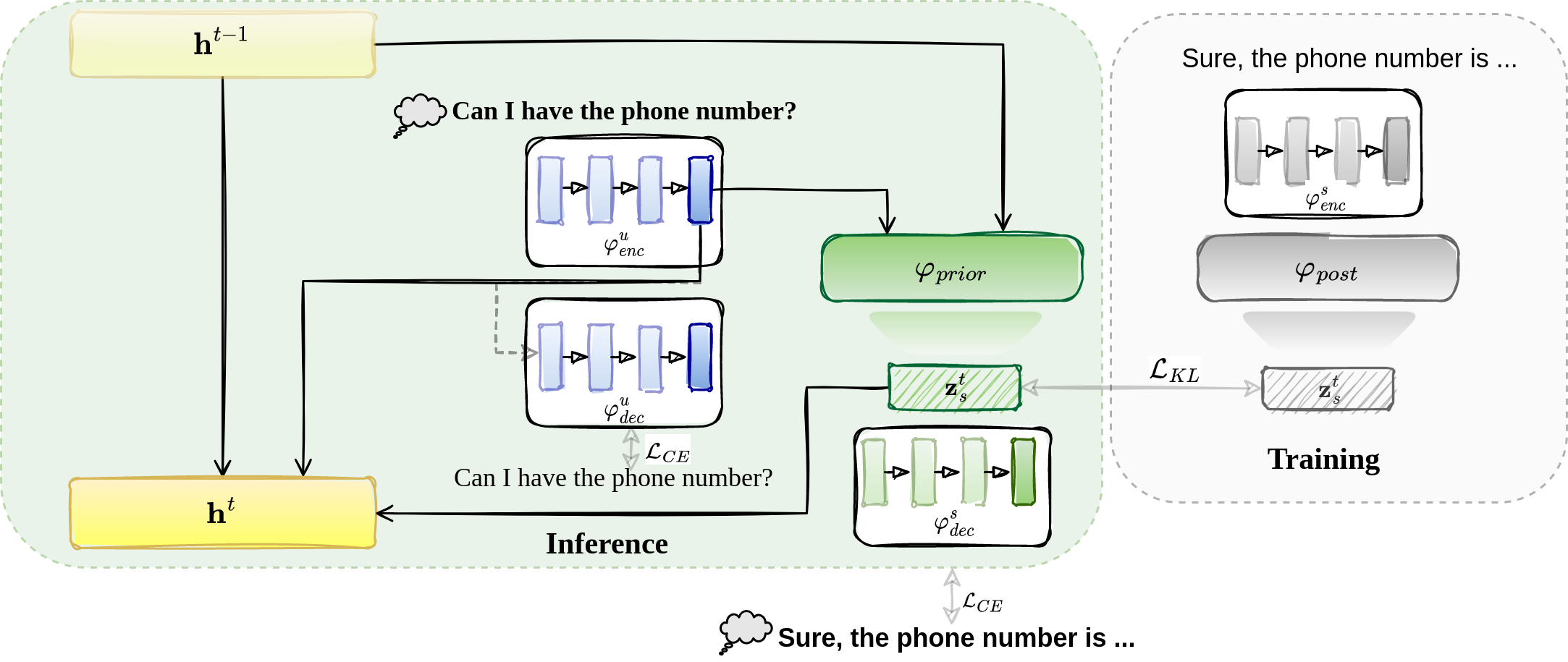}
    \caption{Visualization of our model architecture (one dialogue turn). Yellow boxes represent the turn-level VRNN's hidden state $h^t$. The user utterance is represented as the last hidden state of the encoder network $\varphi_{enc}^u$, which is trained as an autoencoder along with the decoder $\varphi_{dec}^u$. The system utterance, encoded by the network $\varphi_{enc}^s$, is an input to the posterior network $\varphi_{post}$ that helps to train the prior network $\varphi_{prior}$ to construct meaningful latent variables $\mathbf{z}_s$, which initialize the system utterance decoder $\varphi_{dec}^s$. Training uses the whole architecture, including the posterior network $\varphi_{post}$, while only uses the part shaded in green is used for inference.} $\mathcal{L}_{CE}$ stands for cross-entropy loss, $\mathcal{L}_{KL}$ for KL-divergence loss.
    \label{fig:method}
\end{figure*}
most research on unlabeled data only applies to a chit-chat setting. 

Since interpretability and the ability to learn from unlabeled data are our primary goals, we choose working with RNN-based latent variable models over Transformer-based PLMs in our work. 

Unlike previous latent-va\-ria\-ble approaches,
we shift the focus towards task-oriented systems and take tracked entity values and database access into account.
Specifically, we base our approach on \citet{shi2019unsupervised}'s architecture.
\citet{shi2019unsupervised} employ the VRNN model \cite{chung2015recurrent} and experiment with conditioning the prior distribution.
However, their focus is on uncovering dialogue structure, and they model user and system utterances together.
In contrast, we fully take advantage of the VRNN model's generative capabilities and apply it for response generation.
Specifically, we train a specialized decoder for system response generation.
Furthermore, we extend the VRNN model so that the system and the user utterances are modeled separately.
This modification brings the following major advantages:
(1)  We can model different behaviors on the side of the system and the user, which is expected in a task-oriented setting; (2) We can focus on modeling latent system actions in an explainable way; (3) We can predict the system response easily.

Task-oriented dialogue systems typically need to interact with an external database; otherwise, their responses cannot be grounded.
Therefore, we assume that database queries and results are known, but no dialogue state annotation is available.
This allows a direct application of our model for dialogue response generation in a task-oriented setting while still keeping the amount of needed supervision very low.
This scenario reflects the intended use case, i.e.\ automating a call center based on
recordings of previous human-human dialogues.
At some point of the dialogue, a database query is performed by the human agent and we know exactly when and with which parameters.

Our contributions in this paper are as follows:
\begin{enumerate}[itemsep=0pt,topsep=1pt,leftmargin=12pt]
    \item We propose a novel modification of the VRNN-based model for minimally supervised task-oriented dialogue generation, with interpretable latent variables to represent system actions. 
    
    \item We evaluate the system performance in a full task-oriented setting including the database interaction, going beyond previous works in this family of models. Our approach outperforms strong baselines in terms of BLEU and perplexity on three datasets and compares favorably to other baselines.
    \item We present a straightforward way of interpreting the latent variables using a decision tree model.
    We show that our model's latent variables explain most of our system's predicted responses and align well with gold-standard responses.
\end{enumerate}
Our experimental code is released on GitHub.\footnote{\url{https://github.com/vojtsek/to-vrnn}}

\section{Related Work}

In the area of supervised dialogue systems, current leading research focuses on end-to-end sequence-to-sequence models \cite{lei2018}.
Recent works make use of large pre-trained language models (PLMs) 
based on the transformer architecture \cite{vaswani2017} such as GPT-2 \cite{radford2019} or BERT \cite{devlin2019}. 
For example, \citet{wu2020-todbert} 
propose finetuning BERT \cite{devlin2019} for task-oriented dialogue on multiple datasets; 
\citet{zhang2020dialogpt} extended the GPT-2 PLM to model open-domain chit-chat.

However, we focus mainly on approaches that require less supervision.
The hierarchical recurrent encoder-decoder (HRED) by \citet{serban2015building}, where RNN hidden states represent the latent dialogue state, was among the first unsupervised neural dialogue models.
However, the  latent representations obtained from the vanilla autoencoder model trained with reconstruction loss suffer from poor generalization.
For this purpose \cite{bowman-etal-2016-generating}, the usage of Variational Autoencoders (VAEs) \cite{kingma2013auto} was proposed.
The VAE training maximizes the variational lower bound of data log-likelihood.
VAE distributions are invariant in time, therefore it are not suitable for modeling sequences.
\citet{chung2015recurrent} address this issue with the Variational Recurrent Neural Network model (VRNN).
\citet{serban2016hierarchical} then used VRNN's latent variables to represent dialogue state.
Recent works used modified Transformer architectures with specific training tasks to obtain in-context representations of dialogue utterances \cite{bao-etal-2020-plato, liu-etal-2021-dialoguecse}.

While both VAEs and Transformers improve generalization and consistency of the latent variables, they are not well interpretable.
To obtain more interpretable latent states, generative models with discrete states such as hidden Markov models were applied \cite{zhai-williams-2014-discovering, brychcin2016unsupervised}.
\citet{wen2017latent} used discrete latent variables to represent the state in a model trained using reinforcement learning.
Another proposed approach was the usage of quantization techniques by \citet{gunasekara2018quantized},
who perform clustering on utterances and model the dialogue as a sequence of clusters to predict future responses.
\citet{zhao2018unsupervised} use VAEs in combination with Gumbel-Softmax to model discrete latent variables representing the dialogue utterances.

More recently, several works attempted to model latent system actions  without any action-level annotation \cite{huang-etal-2020-generalizable, zhao-etal-2019-rethinking,lubis-etal-2020-lava,zhang_probabilistic_2020}. However, they still rely on labeled data on different levels, such as turn-level dialogue state annotation.
In a different line of research, \citet{shi2019unsupervised} aim to uncover the dialogue structure. They apply VRNNs to estimate dialogue state transition probabilities. The same goal of uncovering and understanding semantic structure of the dialogue is explored by \citet{qiu-etal-2020-structured}, who propose a VRNN-based model with structured attention to achieve this goal, or \citet{sun2021unsupervised}, who use an enhanced graph autoencoder.
Our proposed model combines the latter two approaches, but it is distinct from both.
It models system actions using latent variables, but it does not rely on any turn-level labels for dialogue state or language understanding.
Moreover, our goal is not only to uncover the dialogue structure but rather to model system actions and generate responses.

\begin{table*}[t]
    \centering\scriptsize
    \begin{tabular}{crp{0.75\linewidth}}
        \toprule
         \multirow{4}{*}{\bf Turn 1}&\bf user:& \texttt{Is there a \textbf{moderately priced} restaurant serving \textbf{italian} food anywhere in town?} \\
         &\bf system gold: & \texttt{query italian moderate} \\
         &\bf action: & \texttt{QUERY()} \\
         &\bf system hyp: & \texttt{query \textbf{\textcolor{darkgreen}{italian moderate}}}  \\
         \midrule
        \multirow{4}{*}{\bf Turn 2}&\bf user/database:& \texttt{pizza express, Regent Street City Centre, 01223 324033, C.B 2, 1 D.B, centre}  \\
         &\bf system gold:& \texttt{Pizza express serves italian food and is located in the town centre and is in the moderate price range .} \\        
        &\bf action: & \texttt{OFFER()} \\
        &\bf system hyp:& \texttt{\textbf{\textcolor{red}{Pizza hut Cherry Hinton}} is a italian restaurant in the \textbf{\textcolor{darkgreen}{centre}} part of town} \\
         \midrule
        \multirow{4}{*}{\bf Turn 3}&\bf user:& \texttt{what is the \textbf{address} and \textbf{phone number} ?} \\
        &\bf system gold:& \texttt{their address is  Regent Street City Centre. their phone number is 01223 324033. can i help with anything else ?} \\
         &\bf action: & \texttt{GIVE\_DETAILS()} \\
        & \bf system hyp:& \texttt{the phone number is \textbf{\textcolor{darkgreen}{01223 324033}}. There anything else i can help you with ?} \\
        \bottomrule
    \end{tabular}
    \vspace{-2mm}
    \caption{An example dialogue drawn from the CamRest676 validation set, illustrating the use of database information. We show the user input (or inserted database results), the gold-standard system response, system action annotation based on manual rules (cf.~Section~\ref{sec:pred_latents}), and a prediction of our system (Ours-attn configuration using the database, cf.~Table~\ref{tab:automatic_metrics}). In the first turn, a database query is constructed, the second turn illustrates how the result is retrieved and fed as input. Values inferred correctly by our system are depicted in green, wrong inference is in red.}
    \label{tab:example}
\end{table*}

\section{Method}
We assume that
each dialogue turn $t$ consists of a user utterance $\mathbf{x}_u^t$ and a system utterance $\mathbf{x}_s^t$.
The context $\mathbf{c}^t$ in turn $t$ is a sequence of user and system utterances up to the previous turn $t-1$. 
We expect that conditioning the generation of $\mathbf{x}^t_s$ on a latent variable $\mathbf{z}^t$ will allow the model to better incorporate context.

\subsection{Background: VRNN}
The VRNN model \cite{chung2015recurrent} can be seen intuitively as a recurrent network with a VAE in every timestep.
It extends the VAE model to a sequence of observations generated from a series of hidden latent variables $\mathbf{z}$.
Formally, we want to estimate the joint probability distribution of a sequence of observed and corresponding latent variables $p(\mathbf{x}, \mathbf{z}) = p(\mathbf{x}|\mathbf{z})p(\mathbf{z})$.
The conditional distribution $p(\mathbf{x}|\mathbf{z})$ is parameterized with a neural network.
However, we still need to estimate the posterior $p(\mathbf{z}|\mathbf{x})$ in order to connect the latent variables with the observations.
The VAE uses a variational approximation $q(\mathbf{z}|\mathbf{x})$ that allows to maximize the lower bound of log-likelihood of the data:
\begin{equation}
\begin{split}
    \log~p(\mathbf{x}) \ge -\mathrm{KL}(q(\mathbf{z}|\mathbf{x})||p(\mathbf{z}))\\ + \mathbb{E}_{q(\mathbf{z}|\mathbf{x})}[\log~p(\mathbf{x}|\mathbf{z})]
    \label{eq:vae}
\end{split}
\end{equation}
where KL is the Kullback-Leibler divergence.
We consider a prior  network $\varphi_{prior}$ and a posterior network $\varphi_{post}$, which compute the parameters of $p(\mathbf{z})$ and $q(\mathbf{z}|\mathbf{x})$ respectively.
In a VRNN, $\varphi_{prior}$ and $\varphi_{post}$ additionally depend on the RNN hidden state $\mathbf{h}^t$ to allow for a context-aware prior distribution.
In each time step, we obtain the distribution parameters as follows:
\begin{equation}
\label{eq:distr_theta}
\begin{gathered}
    \mathbf{\theta}_{q}~=~\varphi_{post}(\mathbf{h}^t, \varphi_{enc}(\mathbf{x}^t))\\
    \mathbf{\theta}_{p}~=~\varphi_{prior}(\mathbf{h}^t)
\end{gathered}
\end{equation}
where $\varphi_{enc}$ is the encoder and $\theta_q, \theta_p$ are parameters of the respective distributions (see Section \ref{sec:method_latent}).
With distribution parameters available, we can sample the latent variable and predict the output:
\begin{equation}
\label{eq:x_infer}
\begin{gathered}
    \mathbf{z}^t~\mathtt{\sim}~p(\mathbf{z};\theta_{p})\\
    \mathbf{x}^t~=~ \varphi_{dec}(\mathbf{z}^t)
\end{gathered}
\end{equation}
where $\varphi_{dec}$ represents the decoder network.
The update of the hidden state $\mathbf{h}^t$ is as follows:
\begin{equation}
    \label{eq:hidden_update}
    \begin{gathered}
        \mathbf{h}^{t+1} = \text{RNN}([\varphi_{enc}(\mathbf{x}^t),\varphi_{z}(\mathbf{z}^t)], \mathbf{h}^t)
    \end{gathered}
\end{equation}
where $[.,.]$ is concatenation, $\varphi_z(.)$ is a feature extractor and $\text{RNN}()$ is a step transition function of a recurrent neural network, in our case an LSTM \cite{hochreiter1997}.

\subsection{Modeling task-oriented Dialogue}
We use the VRNN model and extend it to fit the task-oriented setup.
Our model's architecture is depicted in Figure \ref{fig:method}.
We employ a turn-level RNN that summarizes the context to its hidden state.
In each dialogue turn, we model user and system utterances with separate autoencoders to account for different user and system behavior.
The user utterance is modeled with a standard autoencoder; the last encoder hidden state $\varphi^u_{enc}(\mathbf{x}^t_u)$ provides the encoded representation.
For the system part, we use a VAE with discrete latent variables $\textbf{z}_s$ conditioned on the context RNN's hidden state $\mathbf{h}^{t-1}$ and the user utterance encoding $\varphi^u_{enc}(\mathbf{x}^t_u)$.
Our model can thus be seen as a VRNN extended by an additional encoder-decoder module.
The context RNN hidden state update looks as follows:
\begin{equation}
    \begin{gathered}
        \mathbf{h}^{t+1} = \text{RNN}([\varphi^u_{enc}(\mathbf{x}_u^t),\varphi_{z}(\mathbf{z}^t_s)], \mathbf{h}^t)
    \end{gathered}
\end{equation}
For word-level encoding and decoding modules ($\varphi_{enc}^u,\varphi_{enc}^s,\varphi_{dec}^u,\varphi_{dec}^s$), we use an RNN with LSTM cells.
We further experiment with attention \cite{bahdanau2014neural} over user encoder hidden states in the system decoder.
We train the model by minimizing a sum of the cross-entropy reconstruction loss on user utterances and the variational lower bound loss (Equation~\ref{eq:vae}) on system responses.

When running in inference mode, only the prior distribution $p(\mathbf{z}_s)$ is considered, which does not require the system utterance on the input.
Therefore, the model is able to generate the system response when provided with a user utterance on the input.


\subsection{Database interaction}
Task-oriented dialogue systems must provide accurate and complete information based on user requests, which requires external database interaction.
%
To support database access while avoiding costly turn-level annotation,
we follow \citet{bordes2016learning} and 
insert sparse database queries and results directly into the training data, forming special dialogue turns.
Specifically, we identify turns that require database results, e.g.\ to inform about entity attributes or a number of matching entities, and insert a query-result pair in front of those turns (see Table~\ref{tab:example}).We argue that this is the minimal level of supervision required to successfully operate a task-oriented system with database access; it is significantly lower than the full dialogue-state supervision used by most systems.
In addition, it is easily available in the wild (e.g., call center transaction logs).
In practice, we observe that database queries are only inserted for 24\% turns\footnote{This is the average over all datasets in our experiments (see Section~\ref{sec:data}). Per-dataset query counts are 36\%, 23\% and 11\% for CamRest676, MultiWOZ and SMD respectively.} on average.
Note that this approach still covers the task of an explicit state tracker since the necessary entity values are provided when needed.
To maintain consistency, database query results can be stored and used in follow-up questions.

Some experimental approaches, such as \citet{raghu2021unsupervised}, learn database queries without annotation via reinforcement learning. Our framework could use this to handle database interaction more effectively.
We leave this extension for future work.

\subsection{Latent Variables}
\label{sec:method_latent}

We use a set of $n$ \mbox{$K$-way} $(K=20;n=1,3,5)$ categorical variables to achieve good interpretability, following \citet{zhao2018unsupervised}.
This means that each variable is represented as a one-hot vector of length $K$, and we use $n$ such vectors.
We use the Gumbel-Softmax distribution and the reparameterization trick \cite{jang2017categorical}.
During inference, we apply argmax directly to the predicted distribution, instead of sampling from it.
\begin{table}[t]
    \centering\footnotesize
    \begin{tabular}{lrrrrr}
        \toprule
        \textbf{Data} & \textbf{Domains} & \textbf{Slots} & \textbf{Dialogues} & \textbf{T/D}\\ \midrule
        \textbf{MultiWOZ} & 7 & 29 & 10,437 & 13.71 \\
        \textbf{SMD} & 3 & 15 & 3031 & 5.25 \\ 
        \textbf{CamRest676} & 1 & 7 & 676 & 8.12 \\ \bottomrule
        
    \end{tabular}
    \caption{Details of the used datasets giving number of domains, slots, dialogues and average number of turns per dialogue.}
    \label{tab:datasets}
\end{table}

\section{Experiments}

\begin{table*}[t]
    \centering\small
    \begin{tabular}{l|c|rrrr|rrr|rrrr}
      \toprule
      model &  & \multicolumn{4}{c|}{\textbf{CamRest676}} & \multicolumn{3}{c|}{\textbf{SMD}} & \multicolumn{4}{c}{\textbf{MultiWOZ~2.1}} \\
      & db & BLEU & Ppl & MI & EMR & BLEU & Ppl & MI & BLEU & Ppl & MI & EMR \\
    LSTM & \textcolor{red}{\xmark} & 3.90 & 5.34 & -- & -- & 1.62 & 7.84 & -- & 0.92 & 8.23 &
    -- & --\\
    Transformer & \textcolor{red}{\xmark} & 4.98 & 7.72 & -- & -- & 1.53 & 6.33 & -- & 0.95 & 6.95 & -- & -- \\
    GPT-2 & \textcolor{red}{\xmark} & 15.40  & 1.18 & -- & -- & 9.26 & 2.46 & -- & 9.40 & 2.77 & -- & -- \\
    GPT-2 & \textcolor{green}{\cmark} & 13.89 & 1.80 & -- & -- & 4.54 & 2.02 & -- & 9.56 & 2.43 & -- & -- \\
    HRED & \textcolor{red}{\xmark} & \pz2.70 & 13.92 & -- & 0.02 & \pz1.25 & 12.50 & -- & \pz2.98 & 29.61 &
    -- & 0.01\\
    VHRED & \textcolor{red}{\xmark} & \pz4.34 & 11.76 & 0.21 & 0.02 & \pz3.75 & 11.94 & 0.20 & \pz4.65 & 32.74 & 0.15 & 0.01 \\
    VHRED & \textcolor{green}{\cmark} & \pz8.50 & 10.23 & 0.17 & 0.36 & 3.94 & 11.86 & 0.19 & 3.82 & 16.61 & 0.07 & 0.04 \\
    \hdashline[0.5pt/2pt]
    Ours-noattn & \textcolor{red}{\xmark} & 12.98 & \pz4.64 & 0.29 & 0.01 & \pz7.35 & \pz6.18 & \bf0.53 & \pz7.18 & \pz9.16 & \bf0.42 & 0.02\\
    Ours-noattn & \textcolor{green}{\cmark} & 15.10 & \pz4.45 & \bf0.34 & 0.24 & \pz9.24 & \pz\bf6.01 & 0.47 & 11.3 & \pz5.17 & 0.27 & 0.05\\
    Ours-attn & \textcolor{red}{\xmark} & \bf17.37 &\pz 5.07 & 0.16 &  0.09 & 12.30 & \pz6.36 & 0.04 & \bf12.28 & 10.19 & 0.06 & 0.04\\
    Ours-attn & \textcolor{green}{\cmark} & 17.10 & \pz\bf4.23 & 0.22 & \bf0.81 & \bf12.40 & 6.11 & 0.11 & 11.86 & \bf\pz6.03 & 0.05 & \bf0.08\\
    \hdashline[0.5pt/2pt]
    \emph{supervised SotA$^{*}$} & \textcolor{green}{\cmark} & 25.50  & -- & -- & -- & 14.40 & -- & -- & 19.40 & 2.50 & -- & -- \\
    \bottomrule
  \end{tabular}
  \caption{Model performance in terms of Entity Match Rate, BLEU for generated responses, Perplexity (Ppl), and Mutual Information (MI) between the generated response and the latent variables $\mathbf{z}_s$. 
  We measure MI only for the models that use latent variables explicitly. The \emph{db} column indicates systems which use database information. $^{*}$Note that the supervised state-of-the-art scores are not directly comparable, as the systems use full turn-level supervision. Systems listed: CamRest676 \cite{peng2021soloist}; SMD \cite{qin2020dynamic}; MultiWOZ \cite{lin2020mintl}.}
  \label{tab:automatic_metrics}
\end{table*}

\begin{table}[t]
    \centering\small
    \begin{tabular}{l|c|cc}
      \toprule
      config & \textbf{CamRest676} & \multicolumn{2}{c}{\textbf{MultiWOZ 2.1.}} \\
       & gold & domain & action \\
      \midrule
      random  & 0.167 & 0.143 & 0.093 \\
      majority &  0.417 & 0.327 & 0.316 \\\hdashline[0.5pt/2pt]
      HRED & 0.645 & 0.445 & 0.437 \\
      VHRED & 0.521 & 0.357 & 0.323 \\
      GPT-2 & 0.650 & 0.601 & 0.552 \\
      Ours-attn & 0.616 & 0.683 & 0.664 \\
      Ours-noattn & \textbf{0.753} &  \textbf{0.704} & \textbf{0.691} \\\hdashline[0.5pt/2pt]
      Ours-manual & 0.587 & -- & -- \\

      \bottomrule
  \end{tabular}
  \caption{Accuracy of the domain and action decision-tree classifiers based on latent variables. 
  For details about the manual annotation process, see Section~\ref{sec:manual}.}
  \label{tab:latent_classification}
\end{table}

In this section, we focus on the quality of responses generated by our model as well as on model performance with respect to dialogue task success.
We focus on theoretical modeling and feasibility at this stage, which we believe is sufficiently demonstrated by corpus-based evaluation complemented by manual checks. Detailed interpretation of the learned representations follows in Section~\ref{sec:latents}.

\subsection{Data}
\label{sec:data}

We evaluate the model performance on three datasets: CamRest676 \cite{wen2016network}, MultiWOZ 2.1 \cite{budzianowski2018multiwoz,eric2019multiwoz} and Stanford Multidomain Dialogues \cite[SMD;][]{eric-etal-2017-key}\footnote{We use standard splits for MultiWOZ 2.1 and SMD. We split CamRest676 in the 8:1:1 ratio, following previous work.}
All the datasets are task-oriented, i.e., they distinguish between user and system conversational roles.
Furthermore, MultiWOZ and SMD include multiple conversation domains.
The MultiWOZ dataset contains conversations between tourists and a system that provides information about the city they visit, e.g., restaurants, hotels or attractions and transit connections.
SMD contains more concise dialogues between a driver and an in-car virtual assistant. CamRest676 contains only restaurant reservations.
Detailed statistics are given in Table \ref{tab:datasets}.

\paragraph{Database queries} To include database information in the dialogues, we first identify all turns in the original datasets where database information is required, using handcrafted rules.\footnote{These rules are very simple and require minimal effort: whenever database results are provided in the data (based on simple pattern matches over system actions), we prepend a database query based on ground-truth state. The assumption is that in a real-world scenario, these queries would naturally be available -- database queries induced by human operators can be logged along with client-operator conversations.}
We then build database query turns based on the respective state annotation (see example in Table~\ref{tab:example}).
Note that database query parameters are the only annotation used to train our models apart from utterance texts; no other dialogue state annotation from the original datasets is used.
\subsection{Experimental Setup}
\label{sec:expe_setup}
We evaluate two versions of our model: one that uses the attention mechanism (\emph{attn}) and one without it (\emph{noattn}).\footnote{The number and size of the variables are set based on a few cursory checks on the training data. Our models use 10 latent variables by default; we discuss the influence of the number of latent variables in Appendix \ref{sec:appendix}.}
Since our approach is the first to be evaluated in a task-oriented setting with this minimal level of supervision, comparing to prior works is difficult. Setups with full dialog state supervision are not comparable and dialog-state metrics are not applicable without the turn-level supervision. Therefore, we compare our models to standard architectures, such as vanilla LSTM or Transformer encoder-decoder, predicting in a sequence-to-sequence fashion using the same amount of supervision as our approach. We also compare to the HRED/VHRED models, which are perhaps the closest prior work to our approach. To put the results into perspective, we also include scores for fully supervised state of the art on our datasets.
However, note that these scores are not directly comparable.
Model parameters are selected by grid search (see Appendix \ref{sec:ap_training}).\footnote{The training is sensitive to some parameters, such-as the \nobreak{Gumbel-softmax} temperature, but otherwise the model trains easily using conventional optimization methods.}

\subsection{Response quality}
To evaluate the quality of individual responses, we compute BLEU score \cite{papineni2002} and perplexity on the test set (see Table~\ref{tab:automatic_metrics}).

Our architecture performs substantially better than (V)HRED, which commonly fails to pick up the necessary knowledge, especially on larger datasets.
The attention-based versions perform better on BLEU, but lose slightly on perplexity.
Comparing HRED and VHRED shows that using the variational approach generally improves the overall performance.
While the GPT-2 PLM outperforms our approach on perplexity, it is worse on BLEU score, despite its huge capacity.

We compare to other relevant related works:
\begin{enumerate}
\item \citet{shi2019unsupervised} do not use their model for response generation, but they report a negative log likelihood of approximately $5.5 \cdot 10^{4}$ when reconstructing the CamRest676 test set. Our \emph{Ours-noattn} model obtained $0.87 \cdot 10^{4}$, which suggests a better fit of the data.\footnote{This comparison is only approximate since the exact data split is not described by \citet{shi2019unsupervised} -- we are only able to use a test set of the same size, not the exact same instances.}
\item \citet{wen2017latent} measure response generation BLEU score on fully delexicalized CamRest676 data. Their best reported result is 24.60, while our model gets 27.23 (30.10 with attention).
\end{enumerate}

Based on manual checks,
our models are able to generate relevant responses in most cases.
As expected, only the models including database turns are able to predict correct entities (cf.~Section~\ref{sec:emr}).
A relatively common error is informing about wrong slots, e.g.\ the model provides a phone number instead of an address or, even more frequently, provides wrong slot values (cf.~Table~\ref{tab:example}).

\begin{table}[t]
    \centering\small
    \begin{tabular}{lcc}
      \toprule
      model &  success & query acc.\hspace{-2mm} \\
      \midrule
      \multicolumn{3}{c}{\textbf{CamRest676}} \\
      \midrule
      VHRED & 0.21 & 0.91 \\
      Ours-noattn & 0.28 & 0.84 \\\hdashline[0.5pt/2pt]
      supervised SotA \cite{peng2021soloist}\hspace{-2mm} & 0.73 & N/A \\
      \midrule
      \multicolumn{3}{c}{\textbf{MultiWOZ}} \\
      \midrule
      Ours-noattn & 0.10 & 0.98 \\\hdashline[0.5pt/2pt]
      supervised SotA \cite{peng2021soloist}\hspace{-2mm} & 0.85 & N/A \\
      \bottomrule
  \end{tabular}
  \caption{Dialogue success and query accuracy comparison for VHRED, \emph{Ours-noattn} using the database and a state-of-the-art supervised system.}
  \label{tab:success}
\end{table}

\subsection{Task-related performance}
\label{sec:succ}
Without dialogue-state supervision, we cannot measure task-oriented metrics such as \emph{inform} rate or \emph{goal accuracy}.
Therefore, we decided to measure dialogue success and entity match rate, which we adjust to the minimally supervised case (details follow). We also measure database query accuracy.

\paragraph{Dialogue success}
The dialogue success or \emph{success rate} reflects the ratio of dialogues in which the system captures all the mentioned slots correctly and provides all the requested information.
We follow previous works \cite{nekvinda2021shades} and report corpus-based success score, as opposed to using a user simulator.
However, measuring success rate without turn-level labels is not straightforward. 
%
We approximate tracking slot values turn-by-turn by checking for correct slot values upon database queries only, and we use this information to measure dialogue success.
Note that this is not equivalent to having state tracking labels available at all turns, but we consider it a reasonable approximation given our limited supervision -- database queries are crucial for presenting the correct entities to the user, which in turn decides the dialogue success.
The generated query attributes directly show the captured slots.

Success rate results are shown in Table~\ref{tab:success}.
Our system is not competitive with a fully supervised model, but outperforms the baselines (VHRED, GPT).
Upon inspection,
we see that the system is often able to recognize correct slots, however, it has difficulties capturing the correct values.
However, the scores  are promising considering the minimal supervision of our training.

\paragraph{Matching database entities}
\label{sec:emr}
To evaluate the accuracy of the offered entities, we measure the Entity Match Rate (EMR), i.e. the ratio of generated responses with correct entities over all responses that mention some entity.
Table~\ref{tab:automatic_metrics} shows  the results.
We observe that the model performance without the database information is poor.
However, including the database information improves the performance substantially, especially in the case of CamRest676 data.
The MultiWOZ data is much more complex -- it contains more slots and multiple domains that can also be combined in an individual dialogue.
Nevertheless, we can still observe an improvement when we include the database queries.
We also note that using attention improves EMR substantially -- the latent variables alone cannot hold all information about particular values (cf.~Section~\ref{sec:pred_latents}).

\paragraph{Database query accuracy}
Further, we evaluate the accuracy of the database querying.
This metric simply measures if the system queries the database at appropriate turns.
The content of the query is not taken into account in this case, as it is already considered in the success rate. On MultiWOZ, we get a near-perfect accuracy, while our approach loses to VHRED on CamRest676 (see Table~\ref{tab:success}).
We hypothesize that this discrepancy can be caused by different dialogue structures among theses two datasets. The dialogues in CamRest676 usually contain just zero or one query during a dialogue, so our model might generate more queries than necessary.

\section{Latent Variable Interpretation}
\label{sec:latents}
We believe that being able to explain and interpret the model behavior is crucial, especially in a setting without full supervision.
Therefore, we design a set of experiments to evaluate the model behavior and investigate whether the model captures salient dialogue features in the latent variables obtained during training on CamRest676 and MultiWOZ.
While it seems that the latent variables are mainly useful for interpretability or structure induction, they are likely also contributing to the performance as smaller latent spaces yield lower performance as we saw in preliminary experiments and show in Appendix B.

\subsection{Clustering the actions}
\label{sec:clustering}
First, we want to assess whether similar variables represent similar actions.
We follow \citet{zhao2018unsupervised} and define utterance clusters according to the latent variables that have been assigned to them by the model.
We then use the homogeneity metric \cite{rosenberg-hirschberg-2007-v} to evaluate the clustering quality with respect to the reference classes determined by manually annotated system actions (which are used for evaluation only).
Homogeneity reflects the amount of information provided by the clustering (and by extension, the latent vectors used) and is normalized to the interval [0, 1].
The reason of choosing this metric is that it is independent on the number of labels and their permutations.
We provide the results in Table~\ref{tab:homo}.
The clusters formed on the CamRest676 data are more homogeneous than on MultiWOZ, likely because of the greater dataset complexity in the latter case. 
In all cases, our clusters are much more homogeneous than clustering formed by random assignment.
We also compare favorably to stronger baseline that is based on clustering of the sentence representations.
Specifically, in this approach we compute sentence representations using a BERT model tuned for sentence representations \cite{reimers-2019-sentence-bert} and then cluster the obtained sentence embeddings using K-means clustering.

\begin{figure*}[t]
    \centering
    \includegraphics[width=0.70\textwidth]{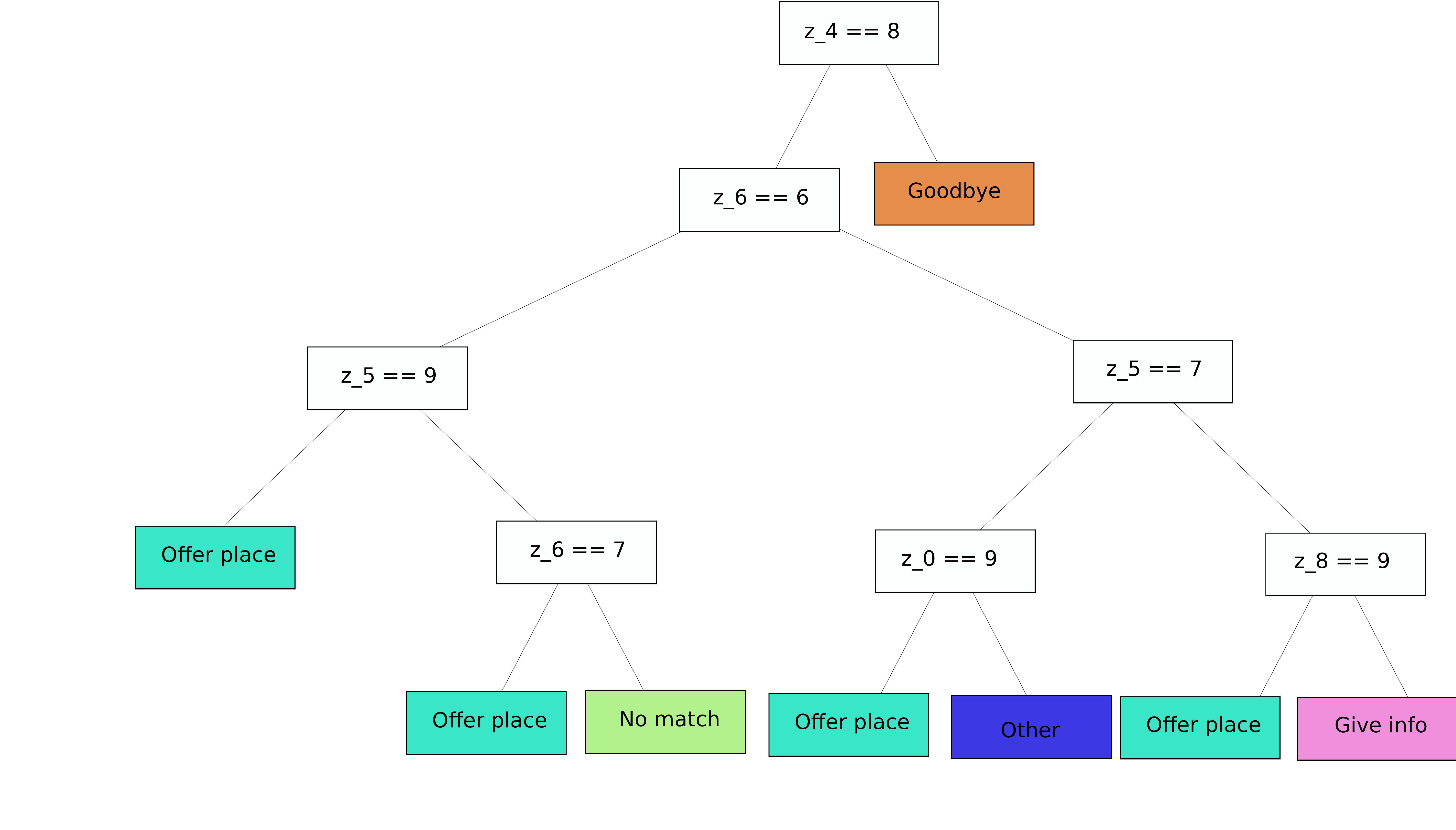}
    \vspace{-5mm}
    \caption{A visualization of a decision tree trained on the CamRest676 data to predict a system action from the contents of the latent variables. Each node represents a decision based on one latent variable value and the leaf node colors represent different system actions. When the condition in a given node is fulfilled, the algorithm proceeds into the right subtree, left otherwise. For clarity, we limit the maximum tree depth to 4. The limit lowers the accuracy slightly -- the pictured tree achieves an accuracy of 73\% on the CamRest676 data.}
    \label{fig:dt}
\end{figure*}

\subsection{Predictive power of the variables}
\label{sec:pred_latents}
To evaluate the predictive power of the obtained latent representations, we train a simple classifier that predicts the system action and current domain, using solely the obtained latent representations as input features.
CamRest676 data does not include system action annotation, hence we manually designed a set of rules to determine system actions.
An example of this rule-based action annotation is shown in Table~\ref{tab:example}.
For MultiWOZ, we predict both system action and the domain of the utterance.

To put our results into perspective, we include several baselines: trivial random and majority class baselines, and classifiers using representations obtained with other methods (HRED, VHRED, GPT).
We use a decision tree (DT) classifier trained with the CART algorithm\footnote{\url{https://scikit-learn.org/stable/modules/tree.html}} and the \emph{gini} split criterion, due to the its good interpretability.
The results are shown in Table~\ref{tab:latent_classification}.
Our classifier beats the random and majority baselines in all cases.
More importantly, it also outperforms classification based on (V)HRED and GPT representations.
This demonstrates that our approach produces high-quality interpretable representations.
We also observe that using attention harms the performance of the action classifier as it makes it possible for the models to bypass the latent variables.

The information about domains and system actions is stored in categorical variables and can be extracted by a simple classification model such as the decision tree which allows us to interpret and explain the behavior of our model.
For illustration, in Figure \ref{fig:dt} we plot a DT with limited depth that achieves 73\% accuracy when predicting the system action on the CamRest676 data.\footnote{The aim is that latent variables hold high-level information, such as intents, actions or domains. This helps interpretability, but is not sufficient for generating appropriate and factually correct responses -- here we need to incorporate correct slot values. This detailed information is captured and carried over via the attention mechanism in \emph{Ours-attn}. Potential alternatives are copy mechanisms \cite{lei2018} or delexicalization on the generated outputs \cite{henderson_robust_2014,peng2021soloist}.}

\subsection{Manual interpretation}
\label{sec:manual}
To explore the interpretability of our representations even further, we manually annotate the latent variables to obtain a simple handcrafted classifier.
Specifically, we draw a set of pairs of utterances and corresponding latent representations from the validation set.
Then we present the representation (discrete) vectors to an expert annotator with a task of assigning an action that each vector represents, based on the sampled utterances.
This way we obtain a mapping from the space of latent vectors to actions.
We then apply this mapping to predict actions on the test set (the \textit{-manual} entry in Table \ref{tab:latent_classification}).
Note that in this approach, we only allow assigning an action to a whole vector, unlike in the case of decision tree classifier that can take individual components into account.
As the results show, this approach works well, despite the above limitation.

\begin{table}[t]
    \centering\small
    \begin{tabular}{l|c|c|c}
      \toprule
      \textbf{Target} & Ours-noattn & sent-repr & random \\
      \midrule
      CamRest676 action & 0.65 & 0.45 & 0.20\\
      MultiWOZ action & 0.34 & 0.33 & 0.02\\
      MultiWOZ domain & 0.39 & 0.30 & 0.01 \\
      \bottomrule
  \end{tabular}
  \caption{Homogeneity for \emph{Ours-noattn} configuration using the database vs.~a clustering of sentence representations and random baseline.}
  \label{tab:homo}
\end{table}

\subsection{Mutual Information}
Finally, we compute mutual information (MI) between the generated text and latent variables as well as among the latent variables themselves (see Table~\ref{tab:automatic_metrics}).\footnote{
Since we measure MI between categorical variables, we quantize the continuous variables used in the VHRED model.}
We see that using attention has a dramatic effect on the amount of MI between the latent variables and the generated text. 
It appears that since attention bypasses the latent vectors, the decoder does not need to use them to store information.

\section{Conclusion and Future Work}

We introduce a model for task-oriented dialogue with discrete latent variables that uses only minimal supervision and improves upon previous approaches \cite{chung2015recurrent,serban2016hierarchical}.
We also propose methods for task-based evaluation in this minimally supervised setting. 
Our system is not yet ready for interactive evaluation on full dialogues, considering the clear 
performance gap with respect to with fully supervised approaches.
However, we demonstrate that it learns meaningful representations from minimal supervision (in a realistic setup corresponding to pre-existing call center call logs) and compares favorably to previous weakly supervised approaches.
A detailed analysis reveals that the learned representations capture relevant dialogue features and can be used to identify system actions.
Furthermore, the reason for choosing an action can be described in an explainable way.
The results suggest that dialogue models with discrete latent variables can be successfully applied also in the task-oriented setting.

The main limitations of our current model are its problems with providing the correct slot values in responses.
We plan address this issue in future work by incorporating explicit copy mechanisms \cite{lei2018}, i.e.\ the model will learn to copy slot values from the context and from database results.
We also plan to experiment with incorporating Transformer models into the variational autoencoder setup, following recent models such as the VAE-transformer \cite{vaeTrans}.

\section{Acknowledgements}
This work was partially supported by  Charles University projects PRIMUS/19/SCI/10, GA UK No.~302120 and SVV No.~260575 and by the European Research Council (Grant agreement No.~101039303 NG-NLG).
It used resources provided by the LINDAT/CLARIAH-CZ Research Infrastructure (Czech Ministry of Education, Youth and Sports project No.~LM2018101).

\bibliography{references}

\begin{thebibliography}{46}
\expandafter\ifx\csname natexlab\endcsname\relax\def\natexlab#1{#1}\fi

\bibitem[{Bahdanau et~al.(2015)Bahdanau, Cho, and Bengio}]{bahdanau2014neural}
Dzmitry Bahdanau, Kyunghyun Cho, and Yoshua Bengio. 2015.
\newblock \href {http://arxiv.org/abs/1409.0473} {Neural {Machine}
  {Translation} by {Jointly} {Learning} to {Align} and {Translate}}.
\newblock In \emph{3rd {International} {Conference} on {Learning}
  {Representations} ({ICLR2015})}, San Diego, CA, USA.

\bibitem[{Bao et~al.(2020)Bao, He, Wang, Wu, and Wang}]{bao-etal-2020-plato}
Siqi Bao, Huang He, Fan Wang, Hua Wu, and Haifeng Wang. 2020.
\newblock \href {https://doi.org/10.18653/v1/2020.acl-main.9} {{PLATO}:
  Pre-trained dialogue generation model with discrete latent variable}.
\newblock In \emph{Proceedings of the 58th Annual Meeting of the Association
  for Computational Linguistics}, pages 85--96, Online. Association for
  Computational Linguistics.

\bibitem[{Bordes et~al.(2017)Bordes, Boureau, and Weston}]{bordes2016learning}
Antoine Bordes, Y{-}Lan Boureau, and Jason Weston. 2017.
\newblock \href {https://openreview.net/forum?id=S1Bb3D5gg} {{Learning}
  {End-to-End} {Goal-Oriented} {Dialog}}.
\newblock In \emph{5th International Conference on Learning Representations,
  {ICLR} 2017}, Toulon, France.

\bibitem[{Bowman et~al.(2016)Bowman, Vilnis, Vinyals, Dai, Jozefowicz, and
  Bengio}]{bowman-etal-2016-generating}
Samuel~R. Bowman, Luke Vilnis, Oriol Vinyals, Andrew Dai, Rafal Jozefowicz, and
  Samy Bengio. 2016.
\newblock \href {https://doi.org/10.18653/v1/K16-1002} {Generating sentences
  from a continuous space}.
\newblock In \emph{Proceedings of The 20th {SIGNLL} Conference on Computational
  Natural Language Learning}, pages 10--21, Berlin, Germany. Association for
  Computational Linguistics.

\bibitem[{Brychcín and Král(2017)}]{brychcin2016unsupervised}
Tomáš Brychcín and Pavel Král. 2017.
\newblock \href {https://doi.org/10.18653/v1/E17-2078} {Unsupervised {Dialogue}
  {Act} {Induction} using {Gaussian} {Mixtures}}.
\newblock In \emph{Proceedings of the 15th {Conference} of the {European}
  {Chapter} of the {Association} for {Computational} {Linguistics}: {Volume} 2,
  {Short} {Papers}}, pages 485--490, Valencia, Spain.

\bibitem[{Budzianowski et~al.(2018)Budzianowski, Wen, Tseng, Casanueva, Ultes,
  Ramadan, and Ga{\v{s}}i{\'c}}]{budzianowski2018multiwoz}
Pawe{\l} Budzianowski, Tsung-Hsien Wen, Bo-Hsiang Tseng, Inigo Casanueva,
  Stefan Ultes, Osman Ramadan, and Milica Ga{\v{s}}i{\'c}. 2018.
\newblock \href {https://www.aclweb.org/anthology/D18-1547} {{MultiWOZ} -- a
  large-scale multi-domain {Wizard-of-Oz} dataset for task-oriented dialogue
  modelling}.
\newblock In \emph{Proceedings of EMNLP}.

\bibitem[{Chung et~al.(2015)Chung, Kastner, Dinh, Goel, Courville, and
  Bengio}]{chung2015recurrent}
Junyoung Chung, Kyle Kastner, Laurent Dinh, Kratarth Goel, Aaron~C Courville,
  and Yoshua Bengio. 2015.
\newblock \href
  {http://papers.nips.cc/paper/5653-a-recurrent-latent-variable-model-for-sequential-data}
  {A recurrent latent variable model for sequential data}.
\newblock In \emph{Advances in neural information processing systems}, pages
  2980--2988.

\bibitem[{Devlin et~al.(2019)Devlin, Chang, Lee, and Toutanova}]{devlin2019}
Jacob Devlin, Ming-Wei Chang, Kenton Lee, and Kristina Toutanova. 2019.
\newblock \href {https://doi.org/10.18653/v1/N19-1423} {{BERT}: Pre-training of
  deep bidirectional transformers for language understanding}.
\newblock In \emph{Proceedings of the 2019 Conference of the North {A}merican
  Chapter of the Association for Computational Linguistics: Human Language
  Technologies (NAACL-HLT)}, pages 4171--4186, Minneapolis, MN, USA.

\bibitem[{Eric et~al.(2020)Eric, Goel, Paul, Sethi, Agarwal, Gao, Kumar, Goyal,
  Ku, and Hakkani-Tur}]{eric2019multiwoz}
Mihail Eric, Rahul Goel, Shachi Paul, Abhishek Sethi, Sanchit Agarwal, Shuyang
  Gao, Adarsh Kumar, Anuj Goyal, Peter Ku, and Dilek Hakkani-Tur. 2020.
\newblock \href {https://www.aclweb.org/anthology/2020.lrec-1.53} {{MultiWOZ}
  2.1: {A} {Consolidated} {Multi}-{Domain} {Dialogue} {Dataset} with {State}
  {Corrections} and {State} {Tracking} {Baselines}}.
\newblock In \emph{Proceedings of the 12th {Language} {Resources} and
  {Evaluation} {Conference}}, pages 422--428, Marseille, France.

\bibitem[{Eric et~al.(2017)Eric, Krishnan, Charette, and
  Manning}]{eric-etal-2017-key}
Mihail Eric, Lakshmi Krishnan, Francois Charette, and Christopher~D. Manning.
  2017.
\newblock \href {https://doi.org/10.18653/v1/W17-5506} {Key-value retrieval
  networks for task-oriented dialogue}.
\newblock In \emph{Proceedings of the 18th Annual {SIG}dial Meeting on
  Discourse and Dialogue}, pages 37--49, Saarbr{\"u}cken, Germany. Association
  for Computational Linguistics.

\bibitem[{Gunasekara et~al.(2017)Gunasekara, Nahamoo, Polymenakos, Ganhotra,
  and Fadnis}]{gunasekara2018quantized}
R~Chulaka Gunasekara, David Nahamoo, Lazaros~C Polymenakos, Jatin Ganhotra, and
  Kshitij~P Fadnis. 2017.
\newblock \href {https://arxiv.org/abs/1812.10356} {Quantized-dialog language
  model for goal-oriented conversational systems}.
\newblock In \emph{DSTC6 -- Dialog System Technology Challenges}, Long Beach,
  CA, USA.
\newblock ArXiv:1812.10356.

\bibitem[{Henderson et~al.(2014)Henderson, Thomson, and
  Young}]{henderson_robust_2014}
Matthew Henderson, Blaise Thomson, and Steve Young. 2014.
\newblock \href {https://doi.org/10.1109/SLT.2014.7078601} {Robust dialog state
  tracking using delexicalised recurrent neural networks and unsupervised
  adaptation}.
\newblock In \emph{2014 {IEEE} {Spoken} {Language} {Technology} {Workshop}
  ({SLT})}, pages 360--365.

\bibitem[{Hochreiter and Schmidhuber(1997)}]{hochreiter1997}
Sepp Hochreiter and J\"{u}rgen Schmidhuber. 1997.
\newblock \href {https://doi.org/10.1162/neco.1997.9.8.1735} {Long short-term
  memory}.
\newblock \emph{Neural Comput.}, 9(8):1735–1780.

\bibitem[{Huang et~al.(2020)Huang, Qi, Sun, and
  Zhang}]{huang-etal-2020-generalizable}
Xinting Huang, Jianzhong Qi, Yu~Sun, and Rui Zhang. 2020.
\newblock \href {https://doi.org/10.18653/v1/2020.findings-emnlp.355}
  {Generalizable and explainable dialogue generation via explicit action
  learning}.
\newblock In \emph{Findings of the Association for Computational Linguistics:
  EMNLP 2020}, pages 3981--3991, Online. Association for Computational
  Linguistics.

\bibitem[{Jang et~al.(2017)Jang, Gu, and Poole}]{jang2017categorical}
Eric Jang, Shixiang Gu, and Ben Poole. 2017.
\newblock \href {https://arxiv.org/abs/1611.01144} {Categorical
  reparameterization with gumbel-softmax}.
\newblock In \emph{International Conference on Learning Representations (ICLR
  2017)}.

\bibitem[{Kingma and Welling(2014)}]{kingma2013auto}
Diederik~P. Kingma and Max Welling. 2014.
\newblock \href {http://arxiv.org/abs/1312.6114} {Auto-{Encoding} {Variational}
  {Bayes}}.
\newblock In \emph{International {Conference} on {Learning} {Representations}
  ({ICLR})}, Banff, AB, Canada.

\bibitem[{Lei et~al.(2018)Lei, Jin, Kan, Ren, He, and Yin}]{lei2018}
Wenqiang Lei, Xisen Jin, Min-Yen Kan, Zhaochun Ren, Xiangnan He, and Dawei Yin.
  2018.
\newblock \href {https://doi.org/10.18653/v1/P18-1133} {{S}equicity:
  Simplifying task-oriented dialogue systems with single sequence-to-sequence
  architectures}.
\newblock In \emph{Proceedings of the 56th Annual Meeting of the Association
  for Computational Linguistics (Volume 1: Long Papers)}, pages 1437--1447,
  Melbourne, Australia.

\bibitem[{Lin et~al.(2019)Lin, Tan, and Frank}]{lin-etal-2019-open}
Yongjie Lin, Yi~Chern Tan, and Robert Frank. 2019.
\newblock \href {https://doi.org/10.18653/v1/W19-4825} {Open sesame: Getting
  inside {BERT}{'}s linguistic knowledge}.
\newblock In \emph{Proceedings of the 2019 ACL Workshop BlackboxNLP: Analyzing
  and Interpreting Neural Networks for NLP}, pages 241--253, Florence, Italy.
  Association for Computational Linguistics.

\bibitem[{Lin et~al.(2020{\natexlab{a}})Lin, Madotto, Winata, and
  Fung}]{lin2020mintl}
Zhaojiang Lin, Andrea Madotto, Genta~Indra Winata, and Pascale Fung.
  2020{\natexlab{a}}.
\newblock \href {https://doi.org/10.18653/v1/2020.emnlp-main.273} {{MinTL:}
  {Minimalist} {Transfer} {Learning} for {Task-Oriented} {Dialogue} {Systems}}.
\newblock In \emph{Proceedings of the 2020 Conference on Empirical Methods in
  Natural Language Processing, {EMNLP} 2020}, pages 3391--3405, Online.

\bibitem[{Lin et~al.(2020{\natexlab{b}})Lin, Winata, Xu, Liu, and
  Fung}]{vaeTrans}
Zhaojiang Lin, Genta~Indra Winata, Peng Xu, Zihan Liu, and Pascale Fung.
  2020{\natexlab{b}}.
\newblock \href {http://arxiv.org/abs/2003.12738} {Variational transformers for
  diverse response generation}.
\newblock \emph{CoRR}, abs/2003.12738.

\bibitem[{Liu et~al.(2021)Liu, Wang, Liu, Sun, Huang, and
  Si}]{liu-etal-2021-dialoguecse}
Che Liu, Rui Wang, Jinghua Liu, Jian Sun, Fei Huang, and Luo Si. 2021.
\newblock \href {https://doi.org/10.18653/v1/2021.emnlp-main.185}
  {{D}ialogue{CSE}: Dialogue-based contrastive learning of sentence
  embeddings}.
\newblock In \emph{Proceedings of the 2021 Conference on Empirical Methods in
  Natural Language Processing}, pages 2396--2406, Online and Punta Cana,
  Dominican Republic. Association for Computational Linguistics.

\bibitem[{Lubis et~al.(2020)Lubis, Geishauser, Heck, Lin, Moresi, van Niekerk,
  and Gasic}]{lubis-etal-2020-lava}
Nurul Lubis, Christian Geishauser, Michael Heck, Hsien-chin Lin, Marco Moresi,
  Carel van Niekerk, and Milica Gasic. 2020.
\newblock \href {https://doi.org/10.18653/v1/2020.coling-main.41} {{LAVA}:
  Latent action spaces via variational auto-encoding for dialogue policy
  optimization}.
\newblock In \emph{Proceedings of the 28th International Conference on
  Computational Linguistics}, pages 465--479, Barcelona, Spain (Online).
  International Committee on Computational Linguistics.

\bibitem[{Nekvinda and Du{\v{s}}ek(2021)}]{nekvinda2021shades}
Tom{\'a}{\v{s}} Nekvinda and Ond{\v{r}}ej Du{\v{s}}ek. 2021.
\newblock \href {https://doi.org/10.18653/v1/2021.gem-1.4} {Shades of {BLEU},
  flavours of success: The case of {M}ulti{WOZ}}.
\newblock In \emph{Proceedings of the 1st Workshop on Natural Language
  Generation, Evaluation, and Metrics (GEM 2021)}, pages 34--46, Online.
  Association for Computational Linguistics.

\bibitem[{Papineni et~al.(2002)Papineni, Roukos, Ward, and Zhu}]{papineni2002}
Kishore Papineni, Salim Roukos, Todd Ward, and Wei-Jing Zhu. 2002.
\newblock \href {https://doi.org/10.3115/1073083.1073135} {{BLEU}: a method for
  automatic evaluation of machine translation}.
\newblock In \emph{Proceedings of the 40th Annual Meeting of the Association
  for Computational Linguistics}, pages 311--318, Philadelphia, Pennsylvania,
  USA.

\bibitem[{Peng et~al.(2021)Peng, Li, Li, Shayandeh, Liden, and
  Gao}]{peng2021soloist}
Baolin Peng, Chunyuan Li, Jinchao Li, Shahin Shayandeh, Lars Liden, and
  Jianfeng Gao. 2021.
\newblock \href {https://doi.org/10.1162/tacl\_a\_00399} {{SOLOIST:} {Building}
  {Task} {Bots} at {Scale} with {Transfer} {Learning} and {Machine}
  {Teaching}}.
\newblock \emph{Trans. Assoc. Comput. Linguistics}, 9:907--824.

\bibitem[{Peng et~al.(2020)Peng, Zhu, Li, Li, Li, Zeng, and Gao}]{peng2020few}
Baolin Peng, Chenguang Zhu, Chunyuan Li, Xiujun Li, Jinchao Li, Michael Zeng,
  and Jianfeng Gao. 2020.
\newblock \href {https://www.aclweb.org/anthology/2020.findings-emnlp.17}
  {Few-shot {Natural} {Language} {Generation} for {Task}-{Oriented} {Dialog}}.
\newblock In \emph{Findings of EMNLP}, pages 172--182.

\bibitem[{Qin et~al.(2020)Qin, Xu, Che, Zhang, and Liu}]{qin2020dynamic}
Libo Qin, Xiao Xu, Wanxiang Che, Yue Zhang, and Ting Liu. 2020.
\newblock \href {https://doi.org/10.18653/v1/2020.acl-main.565} {{Dynamic}
  {Fusion} {Network} for {Multi-Domain} {End-to-end} {Task-Oriented} {Dialog}}.
\newblock In \emph{Proceedings of the 58th Annual Meeting of the Association
  for Computational Linguistics, {ACL} 2020}, pages 6344--6354, Online.

\bibitem[{Qiu et~al.(2020)Qiu, Zhao, Shi, Liang, Shi, Yuan, Yu, and
  Zhu}]{qiu-etal-2020-structured}
Liang Qiu, Yizhou Zhao, Weiyan Shi, Yuan Liang, Feng Shi, Tao Yuan, Zhou Yu,
  and Song-Chun Zhu. 2020.
\newblock \href {https://doi.org/10.18653/v1/2020.emnlp-main.148} {Structured
  attention for unsupervised dialogue structure induction}.
\newblock In \emph{Proceedings of the 2020 Conference on Empirical Methods in
  Natural Language Processing (EMNLP)}, pages 1889--1899, Online. Association
  for Computational Linguistics.

\bibitem[{Radford et~al.(2019)Radford, Wu, Child, Luan, Amodei, and
  Sutskever}]{radford2019}
Alec Radford, Jeffrey Wu, Rewon Child, David Luan, Dario Amodei, and Ilya
  Sutskever. 2019.
\newblock \href {https://openai.com/blog/better-language-models/} {Language
  {Models} are {Unsupervised} {Multitask} {Learners}}.
\newblock Technical report, OpenAI.

\bibitem[{Raghu et~al.(2021)Raghu, Gupta, and Mausam}]{raghu2021unsupervised}
Dinesh Raghu, Nikhil Gupta, and Mausam. 2021.
\newblock \href {https://doi.org/10.1162/tacl\_a\_00372} {{Unsupervised}
  {Learning} of {KB} {Queries} in {Task-Oriented} {Dialogs}}.
\newblock \emph{Trans. Assoc. Comput. Linguistics}, 9:374--390.

\bibitem[{Reimers and Gurevych(2019)}]{reimers-2019-sentence-bert}
Nils Reimers and Iryna Gurevych. 2019.
\newblock \href {https://arxiv.org/abs/1908.10084} {Sentence-bert: Sentence
  embeddings using siamese bert-networks}.
\newblock In \emph{Proceedings of the 2019 Conference on Empirical Methods in
  Natural Language Processing}. Association for Computational Linguistics.

\bibitem[{Rosenberg and Hirschberg(2007)}]{rosenberg-hirschberg-2007-v}
Andrew Rosenberg and Julia Hirschberg. 2007.
\newblock \href {https://www.aclweb.org/anthology/D07-1043} {{V}-measure: A
  conditional entropy-based external cluster evaluation measure}.
\newblock In \emph{Proceedings of the 2007 Joint Conference on Empirical
  Methods in Natural Language Processing and Computational Natural Language
  Learning ({EMNLP}-{C}o{NLL})}, pages 410--420, Prague, Czech Republic.
  Association for Computational Linguistics.

\bibitem[{Serban et~al.(2016)Serban, Sordoni, Bengio, Courville, and
  Pineau}]{serban2015building}
Iulian~V. Serban, Alessandro Sordoni, Yoshua Bengio, Aaron Courville, and
  Joelle Pineau. 2016.
\newblock \href {http://arxiv.org/abs/1507.04808} {Building {End}-{To}-{End}
  {Dialogue} {Systems} {Using} {Generative} {Hierarchical} {Neural} {Network}
  {Models}}.
\newblock In \emph{Proceedings of the 30th {AAAI} {Conference} on {Artificial}
  {Intelligence}}, Phoenix, AZ, USA.

\bibitem[{Serban et~al.(2017)Serban, Sordoni, Lowe, Charlin, Pineau, Courville,
  and Bengio}]{serban2016hierarchical}
Iulian~Vlad Serban, Alessandro Sordoni, Ryan Lowe, Laurent Charlin, Joelle
  Pineau, Aaron Courville, and Yoshua Bengio. 2017.
\newblock \href {http://arxiv.org/abs/1605.06069} {A {Hierarchical} {Latent}
  {Variable} {Encoder}-{Decoder} {Model} for {Generating} {Dialogues}}.
\newblock In \emph{Proceedings of the 31th {AAAI} {Conference} on {Artificial}
  {Intelligence}}, page 3295–3301, San Francisco, CA, USA.

\bibitem[{Shi et~al.(2019)Shi, Zhao, and Yu}]{shi2019unsupervised}
Weiyan Shi, Tiancheng Zhao, and Zhou Yu. 2019.
\newblock \href {https://doi.org/10.18653/v1/N19-1178} {Unsupervised {Dialog}
  {Structure} {Learning}}.
\newblock In \emph{Proceedings of the 2019 {Conference} of the {North}
  {American} {Chapter} of the {Association} for {Computational} {Linguistics}:
  {Human} {Language} {Technologies}, {Volume} 1 ({Long} and {Short} {Papers})},
  pages 1797--1807, Minneapolis, Minnesota.

\bibitem[{Stevens and Su(2021)}]{stevens-su-2021-investigation}
Samuel Stevens and Yu~Su. 2021.
\newblock \href {https://doi.org/10.18653/v1/2021.blackboxnlp-1.34} {An
  investigation of language model interpretability via sentence editing}.
\newblock In \emph{Proceedings of the Fourth BlackboxNLP Workshop on Analyzing
  and Interpreting Neural Networks for NLP}, pages 435--446, Punta Cana,
  Dominican Republic. Association for Computational Linguistics.

\bibitem[{Sun et~al.(2021)Sun, Shan, Tang, Hu, Dai, Yu, Sun, Huang, and
  Si}]{sun2021unsupervised}
Yajing Sun, Yong Shan, Chengguang Tang, Yue Hu, Yinpei Dai, Jing Yu, Jian Sun,
  Fei Huang, and Luo Si. 2021.
\newblock \href {https://ojs.aaai.org/index.php/AAAI/article/view/17634}
  {Unsupervised learning of deterministic dialogue structure with edge-enhanced
  graph auto-encoder}.
\newblock In \emph{Proceedings of the AAAI Conference on Artificial
  Intelligence}, volume~35, pages 13869--13877, Virtual Event.

\bibitem[{Vaswani et~al.(2017)Vaswani, Shazeer, Parmar, Uszkoreit, Jones,
  Gomez, Kaiser, and Polosukhin}]{vaswani2017}
Ashish Vaswani, Noam Shazeer, Niki Parmar, Jakob Uszkoreit, Llion Jones,
  Aidan~N Gomez, {\L}ukasz Kaiser, and Illia Polosukhin. 2017.
\newblock \href {http://arxiv.org/abs/1706.03762} {Attention is all you need}.
\newblock In \emph{Advances in Neural Information Processing Systems
  (NeurIPS)}, pages 5998--6008, Long Beach, CA, USA.

\bibitem[{Wen et~al.(2017{\natexlab{a}})Wen, Miao, Blunsom, and
  Young}]{wen2017latent}
Tsung-Hsien Wen, Yishu Miao, Phil Blunsom, and Steve Young. 2017{\natexlab{a}}.
\newblock \href {http://arxiv.org/abs/1705.10229} {Latent {Intention}
  {Dialogue} {Models}}.
\newblock In \emph{Proceedings of the 34th {International} {Conference} on
  {Machine} {Learning} ({ICML} 2017)}, Sydney, Australia.

\bibitem[{Wen et~al.(2017{\natexlab{b}})Wen, Vandyke, Mrksić, Gašić,
  Rojas-Barahona, Su, Ultes, and Young}]{wen2016network}
Tsung-Hsien Wen, David Vandyke, Nikola Mrksić, Milica Gašić, Lina~M
  Rojas-Barahona, Pei-Hao Su, Stefan Ultes, and Steve Young.
  2017{\natexlab{b}}.
\newblock \href {https://www.aclweb.org/anthology/E17-1042} {A network-based
  end-to-end trainable task-oriented dialogue system}.
\newblock In \emph{Proceedings of EACL}, pages 438--449.

\bibitem[{Wu et~al.(2020)Wu, Hoi, Socher, and Xiong}]{wu2020-todbert}
Chien-Sheng Wu, Steven Hoi, Richard Socher, and Caiming Xiong. 2020.
\newblock \href {https://aclanthology.org/2020.emnlp-main.66/} {{ToD-BERT}:
  Pre-trained natural language understanding for task-oriented dialogues}.
\newblock In \emph{Proceedings of the 2020 Conference on Empirical Methods in
  Natural Language Processing (EMNLP)}, page 917–929, Online.

\bibitem[{Zhai and Williams(2014)}]{zhai-williams-2014-discovering}
Ke~Zhai and Jason~D. Williams. 2014.
\newblock \href {https://doi.org/10.3115/v1/P14-1004} {Discovering latent
  structure in task-oriented dialogues}.
\newblock In \emph{Proceedings of the 52nd Annual Meeting of the Association
  for Computational Linguistics (Volume 1: Long Papers)}, pages 36--46,
  Baltimore, Maryland.

\bibitem[{Zhang et~al.(2020{\natexlab{a}})Zhang, Ou, Hu, and
  Feng}]{zhang_probabilistic_2020}
Yichi Zhang, Zhijian Ou, Min Hu, and Junlan Feng. 2020{\natexlab{a}}.
\newblock \href {https://www.aclweb.org/anthology/2020.emnlp-main.740} {A
  {Probabilistic} {End}-{To}-{End} {Task}-{Oriented} {Dialog} {Model} with
  {Latent} {Belief} {States} towards {Semi}-{Supervised} {Learning}}.
\newblock In \emph{Proceedings of the 2020 {Conference} on {Empirical}
  {Methods} in {Natural} {Language} {Processing} ({EMNLP})}, pages 9207--9219,
  Online. Association for Computational Linguistics.

\bibitem[{Zhang et~al.(2020{\natexlab{b}})Zhang, Sun, Galley, Chen, Brockett,
  Gao, Gao, Liu, and Dolan}]{zhang2020dialogpt}
Yizhe Zhang, Siqi Sun, Michel Galley, Yen-Chun Chen, Chris Brockett, Xiang Gao,
  Jianfeng Gao, Jingjing Liu, and Bill Dolan. 2020{\natexlab{b}}.
\newblock \href {https://doi.org/10.18653/v1/2020.acl-demos.30} {{DIALOGPT} :
  Large-scale generative pre-training for conversational response generation}.
\newblock In \emph{Proceedings of the 58th Annual Meeting of the Association
  for Computational Linguistics (ACL): System Demonstrations}, pages 270--278,
  Online.

\bibitem[{Zhao et~al.(2018)Zhao, Lee, and Eskenazi}]{zhao2018unsupervised}
Tiancheng Zhao, Kyusong Lee, and Maxine Eskenazi. 2018.
\newblock \href {http://aclweb.org/anthology/P18-1101} {Unsupervised {Discrete}
  {Sentence} {Representation} {Learning} for {Interpretable} {Neural} {Dialog}
  {Generation}}.
\newblock In \emph{Proceedings of the 56th {Annual} {Meeting} of the
  {Association} for {Computational} {Linguistics} ({Volume} 1: {Long}
  {Papers})}, pages 1098--1107, Melbourne, Australia.

\bibitem[{Zhao et~al.(2019)Zhao, Xie, and Eskenazi}]{zhao-etal-2019-rethinking}
Tiancheng Zhao, Kaige Xie, and Maxine Eskenazi. 2019.
\newblock \href {https://doi.org/10.18653/v1/N19-1123} {Rethinking action
  spaces for reinforcement learning in end-to-end dialog agents with latent
  variable models}.
\newblock In \emph{Proceedings of the 2019 Conference of the North {A}merican
  Chapter of the Association for Computational Linguistics: Human Language
  Technologies, Volume 1 (Long and Short Papers)}, pages 1208--1218,
  Minneapolis, Minnesota. Association for Computational Linguistics.

\end{thebibliography}
\bibliographystyle{acl_natbib}

\clearpage
\appendix

\section{Training Parameters}
\label{sec:ap_training}
The model is trained with gradient descent, using ADAM optimizer.
We set the hyperparameters according to the BLEU and perplexity results of a grid search on the development set.
Utterance encoder and decoder hidden sizes are 250, the context-LSTM hidden size is 100.
The latent variables are 20-dimensional vectors, their number differs across experiments and is given in the main text.
For the RNN components, we use a dropout probability of $0.3$.
The total model size is 7,047,529 parameters.
The training time is 3-8 hours using one GPU, depending on dataset.

\section{Performance with Various Numbers of Latent Variables}
\label{sec:appendix}
\begin{table}[h]
    \centering\small
    \begin{tabular}{l|ccc}
      \toprule
      & BLEU & Ppl & MI  \\
    \midrule
    Ours-noattn-1z  & 25.2 & 4.25 & 0.46  \\
    Ours-noattn-3z  & 26.8 & 4.24 & 0.26  \\
    Ours-noattn-5z  & 27.23 & 4.20 & 0.38  \\
    Ours-noattn-12z  & 29.83 & 4.12 & 0.35  \\    

    \bottomrule
  \end{tabular}
  \caption{Evaluation of the model performance with respect to automatic measures of BLEU, Perplexity (Ppl) and Mutual Information (MI) on the CamRest676 data.}
  \label{tab:z_counts}
\end{table}

\section{Limitations and risks}
We consider our work to be mostly fundamental research rather than a practical application.
However, it has certain limitations.
Firstly, the proposed way of including the database results is inflexible and it is hard to incorporate possible API changes.
Also, although we show that the latent actions are possible to interpret and explain, with growing number of actions we likely worsen this possibility to interpret the variables.
Another limitation of our current model is its inability to provide correct entities and slot values.

Another limitation and possible risk is that this system is very hard to control and deploying it in current form could produce undesired behavior.
\end{document}